\pdfoutput=1

\documentclass[11pt]{article}

\usepackage{EMNLP2022}

\usepackage{times}
\usepackage{latexsym}

\usepackage[T1]{fontenc}

\usepackage[utf8]{inputenc}

\usepackage{microtype}

\usepackage{inconsolata}

\usepackage{amsmath,amsfonts}
\usepackage{algorithmic}
\usepackage{algorithm}
\usepackage{array}
\usepackage{textcomp}
\usepackage{stfloats}
\usepackage{url}
\usepackage{verbatim}
\usepackage{graphicx}
\usepackage{cite}
\usepackage{color}
\usepackage{booktabs}
\usepackage{multirow}
\usepackage{makecell}
\usepackage{url}
\usepackage{hyperref}
\usepackage{amssymb}
\usepackage{subfigure}

\title{Multimodal Contrastive Learning via Uni-Modal Coding and Cross-Modal Prediction for Multimodal Sentiment Analysis}

\author{Ronghao Lin \and Haifeng Hu\thanks{~~Corresponding author.} \\
     Sun Yat-sen University, China \\
     \tt linrh7@mail2.sysu.edu.cn, huhaif@mail.sysu.edu.cn\\}
\begin{document}
\maketitle
\begin{abstract}
Multimodal representation learning is a challenging task in which previous work mostly focus on either uni-modality pre-training or cross-modality fusion. In fact, we regard modeling multimodal representation as building a skyscraper, where laying stable foundation and designing the main structure are equally essential. The former is like encoding robust uni-modal representation while the later is like integrating interactive information among different modalities, both of which are critical to learning an effective multimodal representation. Recently, contrastive learning has been successfully applied in representation learning, which can be utilized as the pillar of the skyscraper and benefit the model to extract the most important features contained in the multimodal data. In this paper, we propose a novel framework named MultiModal Contrastive Learning (MMCL) for multimodal representation to capture intra- and inter-modality dynamics simultaneously. Specifically, we devise uni-modal contrastive coding with an efficient uni-modal feature augmentation strategy to filter inherent noise contained in acoustic and visual modality and acquire more robust uni-modality representations. Besides, a pseudo siamese network is presented to predict representation across different modalities, which successfully captures cross-modal dynamics. Moreover, we design two contrastive learning tasks, instance- and sentiment-based contrastive learning, to promote the process of prediction and learn more interactive information related to sentiment. Extensive experiments conducted on two public datasets demonstrate that our method surpasses the state-of-the-art methods.
\end{abstract}

\section{Introduction}
With the surge of user-generated videos, Multimodal Sentiment Analysis (MSA) have become a hot research field, which aims to infer people's sentiment based on multimodal data including text, audio and video \citep{zadeh2017tensor,tsai2019multimodal,tsai2020multimodal,poria2020beneath}. To successfully understand human behaviours and interpret human intents, it is necessary to attain an effective and powerful multimodal representation for the model. However, two major challenges in learning such multimodal representation exist: the accurate extraction of uni-modal features and the heterogeneities across different modalities bring difficulty of modeling cross-modal interaction. 

To acquire powerful uni-modal features, \citet{devlin2019bert} presents a large-scale language model named BERT for textual modality and \citet{wu2022wav2clip} introduce an audio representation learning method for audio modality by distilling from \citet{radford2021learning} which targets at transferable model for visual modality. In MSA, previous methods \citep{yu2021learning,han2021improving} mainly utilize BERT for textual modality while vague feature extractor such as COVAREP \citep{degottex2014covarep} and Facet \citep{iMotions} for acoustic and visual modality, where the inherent noise contain in uni-modal features may still exist. To avoid uni-modal noise interfering downstream sentiment inference task, we design Uni-Modal Contrastive Coding (UMCC) which employs feature cutoff strategy inspired by \citep{shen2020simple} and generates augmentation features to construct contrastive learning task with origin uni-modal representation. As shown in Figure \ref{fig1}, we then obtain robust and efficient representations for acoustic and visual modalities.

To alleviate the impact of modality heterogeneity, previous MSA models propose various modalities fusion methods to learn cross-modality interaction information \citep{hazarika2020misa,rahman2020integrating}. Modality translation is a popular method to explicitly translate source modality to the target one, which directly manipulates the commonalities across modalities \citep{tsai2019multimodal,wu2021text,zhao2021missing}. However, due to the existence of discrepancy modality-specific information and huge modality gap, it is undesirable and extremely difficult to project the representations from different modalities to the same one. Different with these explicit modality translation methods, we propose Cross-Modal Contrastive Prediction (CMCP) composed of a pseudo siamese predictive network and two designed contrastive learning tasks to predict cross-modal representation in an implicitly contrastive way. The predictive representation efficiently capture cross-modality dynamics and concurrently preserve modality-specific features for the modalities.

The novel contributions of our work can be summarized as follows:
\begin{itemize}
\item[1)] We propose a framework named MultiModal Contrastive Learning (MMCL), consisting of Uni-Modal Contrastive Coding (UMCC)  which mitigates the interference of modality inherent noise and learns robust uni-modal representations, and Cross-Modal Contrastive Prediction (CMCP) with a pseudo siamese predictive network which learn commonalities and interactive features across different modalities.
\item[2)] We design two contrastive learning tasks, instance- and sentiment-based contrastive learning, in order to improve the convergence of the predictive network and capture sentiment-related information contained in the multimodal data.
\item[3)] We conduct extensive experiments on two publicly available datasets, and gain superior results to the state-of-the-art MSA models.
\end{itemize}
\section{Related Work}
\subsection{Multimodal Sentiment Analysis (MSA)}
MSA focus on integrating textual, acoustic and visual modalities to comprehend varied human sentiment \citep{morency2011towards}. Previous research mainly comprises of two steps: uni-modal representation learning and  multimodal fusion. For uni-modal representation, \citet{tsai2019learning} factorizes them into two independent sets while \citet{hazarika2020misa} projects them into two distinct subspaces. Large pre-trained Transformer-based language models such as BERT have shown great performance improvement on downstream NLP tasks \citep{devlin2019bert}. However, for acoustic and visual modalities, the features are extracted by CMU-MultimodalSDK with a vague description of feature and backbone selection\citep{zadeh2018multi} in MSA task. We argue that without powerful pre-trained language tokenizer to extract features as textual modality does, the inherent noise of acoustic and visual features may disturb the inference of sentiment.

Different from direct processing the uni-modal features, we present uni-modal features coding to learn robust acoustic and visual representations. For multimodal fusion, \citet{zadeh2017tensor,liu2018efficient,zadeh2018multi} present early fusion at the feature level while \citet{poria2017context,zadeh2018memory,yu2021learning} adopt late fusion at the decision level. However, the former methods limit capabilities in modeling cross-modal dynamics due to the inconsistent space of different modalities while the later methods suffer from neglecting modality-specific information with the absence of low-level feature process. To avoid the respective issues from the two methods, \citet{tsai2019multimodal,tsai2019learning,hazarika2020misa} propose hybrid fusion which perform multimodal fusion at both input and output level. Guided by this thought, we construct a cross-modal predictive network to process representations from different modalities at early and late fusion stage, which effectively exploit the intra- and inter-modality dynamics in a prediction manner.
\subsection{Contrastive Learning}
The core idea of contrastive learning is to measure the similarities of sample pairs in the representation space \citep{hadsell2006dimensionality}, which is firstly adopted in the field of computer vision \citep{he2020momentum} and then extend to the field of nature language analysis \citep{gao2021simcse}. Previous work based on contrastive learning mostly only consider uni-modal data and utilize contrastive losses in a discrimination manner. Different with discrimination models, \citet{van2018representation} combines predicting future observations named predictive coding with a probabilistic contrastive loss called InfoNCE. Inspired by but diverse from this work, we apply predictive network with contrastive learning in multimodal feature to capture cross-modal dynamics and enhance the interaction among different modalities.
\section{Method}
\subsection{Problem Definition}
In MSA task, the input is utterance consisting of three modalities: textual, acoustic and visual modality, where $m\in\{t,a,v\}$ . The sequences of these three modalities are represented as triplet $(T,A,V)$, including $T \in \mathbb{R}^{N_t\times d_t}$, $A \in \mathbb{R}^{N_a\times d_a}$ and $V \in \mathbb{R}^{N_v\times d_v}$ where $N_m$ denotes the sequence length of corresponding modality and $d_m$ denotes the dimensionality. The goal of MSA task is to learn a mapping $f(T,A,V)$ to infer the sentiment score $\hat{y}\in \mathbb{R}$.
\subsection{Overall Architecture}
\begin{figure*}[htbp]
	\centering 
	\includegraphics[scale=0.42]{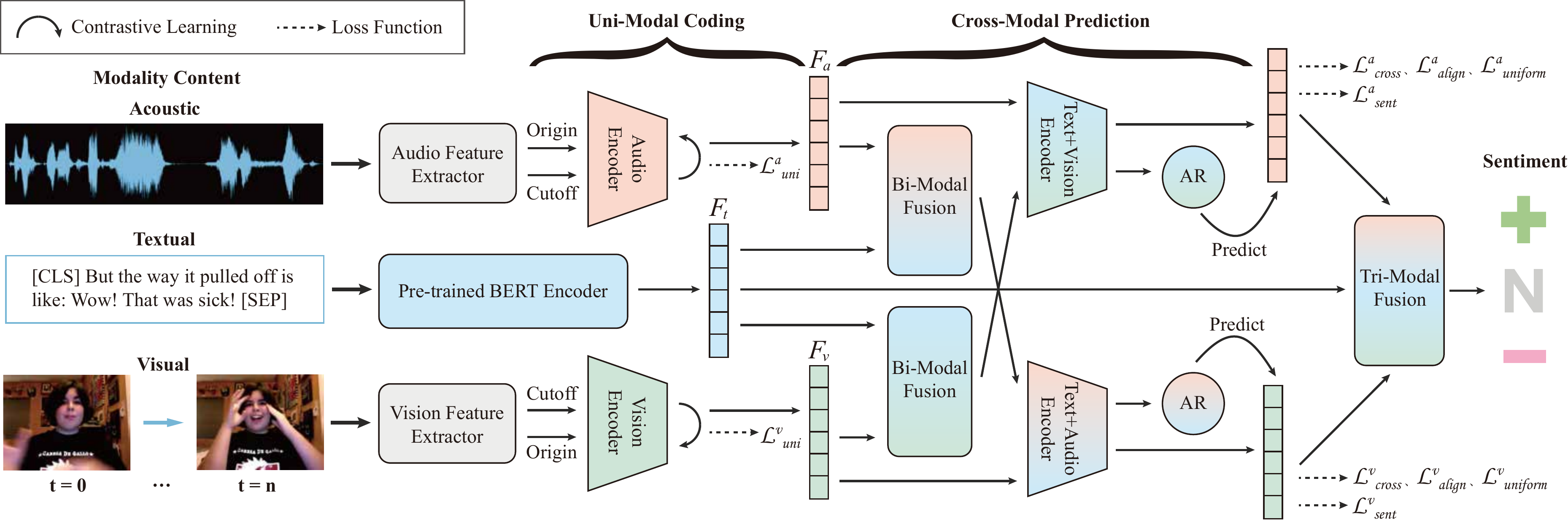}
	\caption{The overall architecture of our proposed MMCL framework.}
	\label{fig1}
\end{figure*}

As shown in Figure \ref{fig1}, we firstly process raw input into sequential feature vectors with fixed feature extractor for audio and vision data while pre-trained BERT \citep{devlin2019bert} encoder for text. Then we utilize contrastive learning in both uni-modal coding and cross-modal prediction, which are the two key modules in our proposed model. The uni-modal coding drive the model to focus on informative features which then implicitly filter out inherent noise and produces robust and effective uni-modal representation for acoustic and visual modalities. The cross-modal prediction captures commonalities among different modalities and outputs predictive representation full of interaction dynamics. Lastly, we fuse predictive acoustic and visual representations with textual representation to derive the final multimodal representation which contains both modality-specific and cross-modal dynamics most related to sentiment.
\subsection{Uni-Modal Contrastive Coding}
For uni-modality, we encode the sequential triplet $(T,A,V)$ into corresponding representations. Specifically, we use BERT \citep{devlin2019bert} to encode input sentences to obtain the hidden representations of textual modality. The embedding from the last Transformer layer's output can be represented as:
\begin{equation}
	F_t = BERT(T;\theta^{BERT}_t)\in\mathbb{R}^{L_t\times d_t}
\end{equation}

To acquire more robust acoustic and visual representations, we design Uni-Modal Contrastive Coding (UMCC) for both modalities. Firstly, we encode audio and vision inputs by two uni-modal bi-directional LSTMs \citep{hochreiter1997long} to capture temporal characteristic:
\begin{equation}
\begin{aligned}
	h_a & = bLSTM(A;\theta^{bLSTM}_a)\in\mathbb{R}^{L_a\times d_a}\\
	h_v & = bLSTM(V;\theta^{bLSTM}_v)\in\mathbb{R}^{L_v\times d_v}
\end{aligned}
\end{equation}

\begin{figure*}[htbp]
	\centering 
	\includegraphics[scale=0.55]{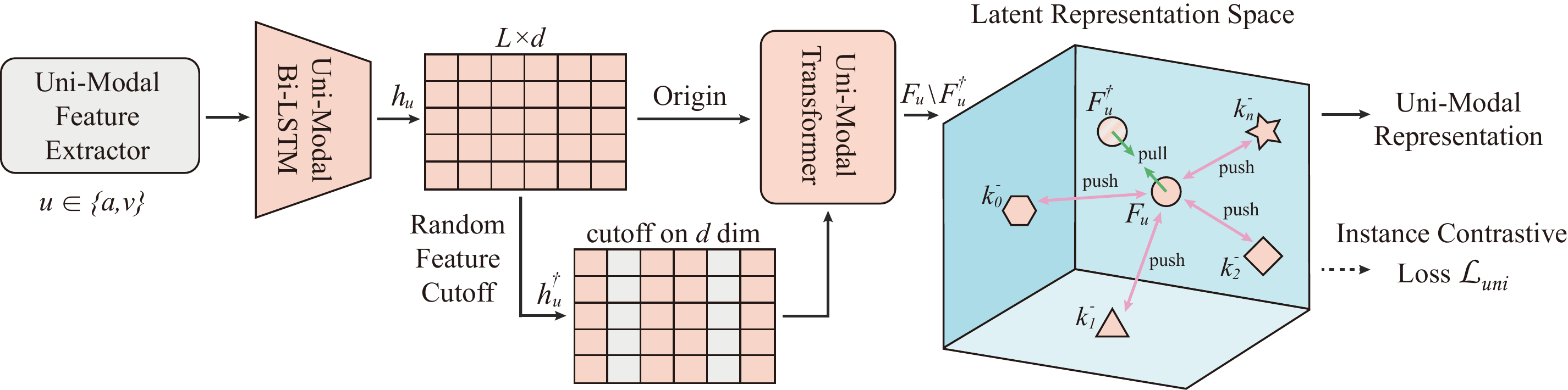}
	\caption{Illustration of Uni-Modal Contrastive Coding.}
	\label{fig2}
\end{figure*}

To construct contrastive learning, we treat the encoded acoustic and visual representations as query samples $q$ and get the corresponding positive key samples $k^+$ by feature augmentation strategy. In natural language understanding and generation task, \citet{shen2020simple} introduces an efficient data augmentation approach named cutoff to erase part of the information within an input sentence and yield its restricted views during the fine-tuning stage. Inspired by this work, we utilize random feature cutoff strategy on acoustic and visual representations which randomly convert a certain proportion of embedding dimensions of every token within the sequence into a vector of zeros. As shown in Figure \ref{fig2}, we then generate augmented version of uni-modal representations, denoted as $h_a^\textit{\dag}$ and $h_v^\textit{\dag}$. 

In order to fuse with textual representations in the similar semantic space later, we design uni-modal Transformer models for acoustic and visual modalities, respectively. For $u\in\{a,v\}$, the query sample $q$ and positive key sample $k^+$ are respectively denoted as $F_u$ and $F^\textit{\dag}_u$:
\begin{equation}
\begin{aligned}
	F_u & = Transformer(h_u;\theta^{Tf}_u)\in\mathbb{R}^{L_u\times d_u}\\
	F^\textit{\dag}_u & = Transformer(h^\textit{\dag}_u;\theta^{Tf}_u)\in\mathbb{R}^{L_u\times d_u}
\end{aligned}
\end{equation}

Given a batch set $F_{uni}=\{F_u^0, F_u^1, ..., F_u^{n-1}\}$, noted that there is a single positive key $F^\textit{\dag}_u$ (as $k^+$) that each encoded query $F_u^i$ (as $q$, $i\in[1,n]$) matches, while the other representations $F_u^j$ ($j\in[0,n]$ and $j\neq i$) in the same batch are considered as negative key samples $k^-$, as the instance discrimination task \citep{wu2018unsupervised, ye2019unsupervised} does. With the similarity measured by dot product, we present the uni-modal instance contrastive loss $\mathcal L_{uni}$ in InfoNCE \citep{van2018representation} form:
\begin{equation}
\begin{aligned}
	\mathcal L^u_{uni} & \triangleq -\log \frac{\exp(q\cdot k^+/\tau)}{\sum_{i=1}^{n}\exp(q\cdot q^i/\tau)}\\
	& = -\mathop{\mathbb{E}}\limits_{F_{uni}}\left[ \log \frac{\exp(F_u\cdot F^\textit{\dag}_u/\tau)}{\sum_{i=1}^{n}\exp(F_u\cdot F_u^i/\tau)}\right]
\end{aligned}
\end{equation}
where $\tau$ is a temperature hyper-parameter that controls the probability distribution over distinct instances \citep{hinton2015distilling}. Due to $u\in\{a,v\}$, the final uni-modal instance contrastive loss $\mathcal L_{uni}=\mathcal L^a_{uni}+\mathcal L^v_{uni}$.

Since each dimension of representation contains certain features, with a certain number of features erased entirely, the UMCC can impel uni-modal Transformer models to implicitly eliminate the inherent noise of uni-modal data and capture informative semantic information which are the most essential to predict the sentiment. Doing so, we can finally acquire a robust modality-specific representation for acoustic and visual modalities.
\subsection{Cross-Modal Contrastive Prediction}
\begin{figure*}[htbp]
	\centering 
	\includegraphics[scale=0.48]{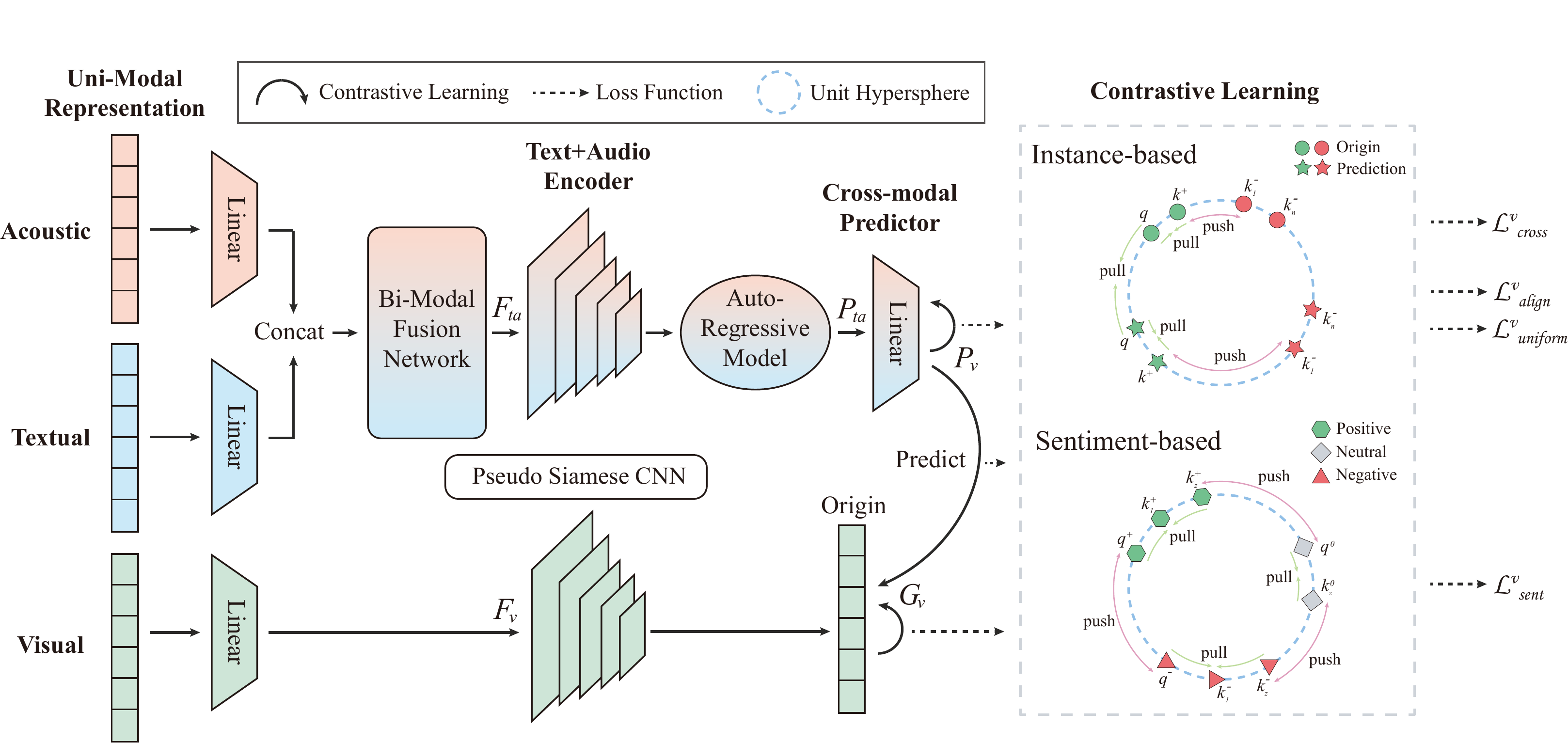}
	\caption{Illustration of Cross-Modal Contrastive Prediction along with two contrastive learning tasks.}
	\label{fig3}
\end{figure*}

 To further learn inter-modality dynamics and focus on the commonalities related to sentiment among different modalities, we present Cross-Modal Contrastive Prediction (CMCP) as shown in Figure \ref{fig3}. Specifically, we construct a pseudo siamese predictive network to utilize two modalities to predict another one. Through the network we can attain the original representation and the corresponding predictive representation. For intra- and inter-embedding of these two representations, we design two contrastive learning tasks based on data instances and sentiment labels, respectively. The experiment in Section \ref{sec_result} indicate the effectiveness of the designed contrastive learning tasks.
 
\subsubsection{Pseudo Siamese Predictive Network}
The proposed pseudo siamese predictive network consists of a bi-modal fusion network, pseudo siamese convolutional neural networks (CNNs), an auto-regressive model and a cross-modal predictor.

As shown in Figure \ref{fig3}, we take acoustic and textual modalities to predict visual modality as an example. Firstly, on account of various feature dimensions of different modalities, we utilize a linear layer to project three representations to the same dimension. 

Secondly, we concatenate acoustic and textual representations and feed them into the bi-modal fusion network, which is a multilayer perceptron (MLP). After that, we can obtain the bi-modal representation denoted as $F_{ta}$, where $F_{ta}=MLP(\{F_t;F_a\})$. To mine out the commonalities among three modalities, the bi-modal representation $F_{ta}$ are utilized to predict the visual modality representation $F_v$ subsequently.

Thirdly, compressing high-dimension embedding into a more compact latent representation space can make conditional predictions easier to the predictive network \citep{van2018representation}. To project $F_{ta}$ and $F_v$ into the same compact space, we design two CNNs encoders in pseudo siamese structure \citep{bromley1993signature}. As shown in Figure \ref{fig3}, the pseudo siamese CNNs means that two CNNs have the same architecture setting but with unshared parameters, which can model the inputs in the same way while with different neuron weights according to their own characteristics. In this compact latent space, visual representation is denoted as $G_v$, which is the prediction target subsequently.

After mapping the bi-modal representations $F_{ta}$ by CNN encoder, an auto-regressive model is applied to summarize and produce a context latent representation $P_{ta}$, where $P_{ta}=sLSTM(F_{ta})$. We regard $P_{ta}$ as an inter-modality dynamics container which covers common features shared by bi-modalities and visual modality. 

At last, we use a linear projection layer as a cross-modal predictor and get the predictive visual representation $P_v$, where $P_{v}=Linear(P_{ta})$. For $u\in\{a,v\}$, Now the problem has become how to make the prediction $P_{u}$ more consistent with the target $G_{u}$ while retaining the most useful information for downstream tasks.
\subsubsection{Instance-based Contrastive Learning}
To maximize the similarity between the prediction representation $P_{u}$ and target representation $G_{u}$, we propose the Instance-based Contrastive Learning (ICL) to force the predictive network learning effective features in the original representation. 

In details, we firstly perform L2-normalization on both $P_{u}$ and $G_{u}$ to restrict the predictive space to the unit hypersphere \citep{wang2020understanding}, as shown in Figure \ref{fig3} with $u=v$ as an example. Considering the instability of prediction training at the beginning, we then successively take the combination of original and prediction representation as query and key, meaning that $\{query,key\}$ can be $\{origin, origin\}$, $\{predict, predict\}$ and $\{origin, predict\}$. The reason of this setting will be further discussed with experiment evidence in Section \ref{sec:loss}. In ICL, each query $q$ has a corresponding key as $k^+$ while the other representations in the same batch are seen as $k^-$. Similar with uni-modal instance contrastive loss $\mathcal L_{uni}$, the cross-modal instance contrastive loss $\mathcal L_{cross}$ is presented as:
\begin{equation}
	\mathcal L_{cross} \triangleq -\mathop{\mathbb{E}}\limits_{F_{cross}}\left[ \log \frac{\exp(F_{c}\cdot F^+_{c}/\tau)}{\sum_{i=1}^{n}\exp(F_c\cdot F_c^{i}/\tau)}\right]
\end{equation}
where $F_c\backslash F_c^+\in\{P_{u},G_{u}\}$, $u\in\{a,v\}$ and $F_{cross}=\{F_c^1, ..., F_c^n\}$. 

Due to the discrepancy modality-specific information contained in different modalities, there exists huge modality gap across the representations of modalities. Intuitively, contrastive learning task on cross-modal prediction is far more difficult than on uni-modal coding. Inspired by \citep{wang2020understanding} which identifies two key properties related to contrastive loss, we introduce two more contrastive loss for better optimization of the predictive cross-modal instance-based contrastive learning:
\begin{itemize}
	\item
	Cross-modal alignment loss $\mathcal L_{align}$ to map query sample $q$ and positive key sample $k^+$ from different modalities to nearby features and thus be mostly invariant to the modality gap, straightforwardly defined with the expected distance among two samples:
	\begin{equation}
	\label{equ6}
	\mathcal L_{align} \triangleq \mathop{\mathbb{E}}[  \Vert P_u - G_u \Vert_2^\lambda],\ \  \lambda > 0
    \end{equation}
	
	\item
	Cross-modal uniformity loss $\mathcal L_{uniform}$ to distribute predictive and target representations roughly uniformly on the unit hypersphere and preserve as much  modality-specific information as possible, defined as the logarithm of the average pairwise Gaussian potential\citep{cohn2007universally}:
	\begin{equation}
	\label{equ7}
	\mathcal L_{uniform} \triangleq \log\mathop{\mathbb{E}}\left[  e^{-\kappa\Vert P_u - G_u \Vert_2^2}\right],    \ \ \kappa > 0
    \end{equation}
\end{itemize}
\subsubsection{Sentiment-based Contrastive Learning}
For the purpose of concentrating on sentiment related features contained in multimodal representations, we construct Sentiment-based Contrastive Learning (SCL) to help the modal learn more discriminative representation for different polarities of sentiments. 

Firstly, according to the sentiment labels, we divide the representations into three sentiment classes $\{positive(+), neutral(0), negative(-)\}$. Then, we treat the representations from the same sentiment class as positive pairs, represented as $\{q^s, k_1^s, ..., k_{z}^s\}$ where $s\in\{+,0,-\}$ and $z\leq n$. The other representations from different sentiment classes in the same batch are treated as negative samples. At last, the sentiment-based contrastive learning loss $\mathcal L_{sent}$ can be given by:
\begin{equation}
	\mathcal L_{sent} \triangleq -\mathop{\mathbb{E}}\limits_{F_{sent}}\left[ \log \frac{\sum_{j=1}^{z}\exp(q_{s}\cdot k^s_{j}/\tau)}{\sum_{i=1}^{n}\exp(q_{s}\cdot k^s_{i}/\tau)}\right]
\end{equation}
where $F_{sent}$ denotes original and corresponding predictive representations divided by sentiment classes. Here we consider both original and predictive representations which is conducive to guide the predictive network to preform hard sample mining when the sentiments across different modalities are variant.

By means of the sentiment-based contrastive learning, we extend contrastive learning approach to a fully-supervised setting as \citep{khosla2020supervised} does and effectively leverages sentiment label information for better downstream sentiment classification performance.
\subsection{Total Training Loss}
At last, we fuse the predictive acoustic and visual representations along with the textual representation to attain the final multimodal representation, represented as $F_M=MLP(\{F_t;P_a;P_v\})$, and predict the final sentiment score $\hat{y}$. Along with the truth sentiment label $y$, we can have the regression task loss $\mathcal L_{reg}$ as:
\begin{equation}
	\mathcal L_{reg} = \frac{1}{n}\sum^{n}_{i=1}|y_i-\hat{y}_i|
\end{equation}
where $n$ is the number of training samples. Combined with the designed contrastive losses, the total loss for training is formulated as:
\begin{equation}
	\hspace{-2.mm}
    \begin{aligned}
	\mathcal L_{total} = & \mathcal L_{reg} + \mu\mathcal L_{uni} + \eta\mathcal L_{sent} \\
	& + \alpha\mathcal L_{cross} +\beta\mathcal L_{align} +\gamma\mathcal L_{uniform}
	\end{aligned}
    \label{equ10}
\end{equation}
where $\mu$, $\eta$, $\alpha$, $\beta$ and $\gamma$ are weighted hyper-parameters that adjust the impact of various loss functions.
\section{Experiments}
We conduct extensive experiments on two public datasets which offer both token-aligned and -unaligned data for multimodal sentiment analysis. The details about the datasets, evaluation metrics and baseline methods are provided subsequently.
\subsection{Datasets and Evaluation Metrics}
\textbf{CMU-MOSI} \citep{zadeh2016mosi} is a popular benchmark dataset collected from YouTube in MSA research, including 93 monologues where speakers make their comment on a specific topics. The dataset consists of 2,199 opinion video segments with a total of 26,295 words in the utterances and is annotated with sentiment intensity label ranged from -3 (strongly negative) to +3 (strongly positive).

\textbf{CMU-MOSEI} \citep{zadeh2018multimodal} is a large dataset of multimodal sentiment analysis and emotion recognition, containing 23,454 YouTube monologues video segments covering 250 distinct topics from 1,000 distinct speakers. The utterances in the dataset are randomly chosen on various movie review topics, annotated with sentiment scores between -3 and +3 and 6 different emotion classes. 

Both of the datasets are split into train, validation and test sets as \citet{han2021improving} does.

We use public evaluation metrics of classification and regression to demonstrate the performance of our proposed framework and further compare with baselines: seven-class classification accuracy (Acc7) indicating the correct sentiment label predictions in the range of [-3, +3], binary classification (Acc2) and F1-score using two calculation settings marked as segmentation symbol '-$/$-' where left represents non-negative$/$negative (has-0) and right denotes positive$/$negative (non-0); mean absolute error (MAE) computing the average absolute difference between predicted and truth labels, Pearson correlation (Corr) measuring the degree of prediction skew.
\begin{table*}[htbp]
	\centering
	\setlength\tabcolsep{2.5pt}
	\begin{tabular}{ccccccccccc}
		\toprule[1.5pt]
		\multirow{2}{*}{Models} & \multicolumn{5}{c}{CMU-MOSI} & \multicolumn{5}{c}{CMU-MOSEI}\\
		\cmidrule(r){2-6}\cmidrule(r){7-11}
		& Acc7$\uparrow$  & Acc2$\uparrow$ & F1$\uparrow$ & MAE$\downarrow$ & Corr$\uparrow$ & Acc7$\uparrow$  & Acc2$\uparrow$ & F1$\uparrow$ & MAE$\downarrow$ & Corr$\uparrow$\\
		\midrule[1.5pt]
		TFN$^*$ & 33.7  &  78.3/80.2  &  78.2/80.1  & 0.925  & 0.662  &  
		52.2 & 81.0/82.6  &  81.1/82.3  & 0.570 & 0.716 \\
		LMF$^*$ & 32.7  &  77.5/80.1  &  77.3/80.0  & 0.931  & 0.670  & 
		52.0 & 81.3/83.7  &  81.6/83.8  & 0.568 & 0.727 \\
		MFN & 34.2 &  77.9/80.0  &  77.8/80.0 & 0.951 & 0.665  & 
		51.1 & 81.8/84.0  &  81.9/83.9  & 0.575 & 0.720 \\
		MFM & 33.3 &  77.7/80.0  &  77.7/80.1 & 0.948 & 0.664  &  
		50.8 & 80.3/83.4  &  80.7/83.4  & 0.580 & 0.722 \\
		MulT & 35.0 &  79.0/80.5  &  79.0/80.5  & 0.918 & 0.685  &
		52.1 & 81.3/84.0  &  81.6/83.9  & 0.564 & 0.732 \\
		MISA & 43.5 &  81.8/83.5  &  81.7/83.5  & 0.752 & 0.784 & 
		52.2 & 81.6/84.3  &  82.0/84.3  & 0.550  & 0.758 \\
		MAG-BERT & 45.1  &  82.4/84.6  &  82.2/84.6  &  0.730 & 0.789 &
		52.8 &  81.9/85.1  &  82.3/85.1  & 0.558  & 0.761 \\
		Self-MM &  45.8 &  82.7/84.9  &  82.6/84.8  & 0.731  & 0.785  & 
		53.0 &  82.6/85.2  &  82.8/85.2  &  0.540 & 0.763 \\
		MMIM & 45.0 &  83.0/85.1  &  82.9/85.0  & 0.738  & 0.781  & 
		53.1 &  81.9/85.1  &  82.3/85.0  & 0.547  &  0.752 \\
		\midrule[0.5pt]
		MMCL(ours) & \textbf{46.5}  &  \textbf{84.0}/\textbf{86.3}  &  \textbf{83.8}/\textbf{86.2}  &  \textbf{0.705} & \textbf{0.797}  & \textbf{53.6}  &  \textbf{84.8}/\textbf{85.9}  &  \textbf{84.8}/\textbf{85.7}  & \textbf{0.537}  &  \textbf{0.765} \\
		\bottomrule[1.5pt]     
	\end{tabular}
	\caption{Performance Comparison between MMCL and baselines on CMU-MOSI and CMU-MOSEI datasets. The multimodal data are token-aligned except for unaligned models with $^*$. }
	\label{table1}
\end{table*} 

\subsection{Baselines}
The mentioned baselines in the experiment are introduced in detail in the following.

\textbf{TFN} \citep{zadeh2017tensor} Tenser Fusion Network introduces a multi-dimensional tensor by calculating the outer-product among different modalities to capture uni-modal, bi-modal and tri-modal interactions.

\textbf{LMF} \citep{liu2018efficient} Low-rank Multimodal Fusion decomposes stacked high-order multimodal tensors into low-rank weight tensors to reduce computational complexity and perform efficient fusion.

\textbf{MFN} \citep{zadeh2018memory} Memory Fusion Network separately leverages LSTM to encodes information from each modality and utilizes a delta-memory attention network with a multi-view gated memory to explicitly accounts for the cross-view interaction.

\textbf{MFM} \citep{tsai2019learning} Multimodal Factorization Model presents jointly optimize multimodal discriminative factors and modality-specific generative factors to reconstruct missing modalities and interpret the interactions that influence multimodal learning.

\textbf{MulT} \citep{tsai2019multimodal} Multimodal Transformer extends three sets of Transformers with directional pairwise cross-modal attention which latently adapts streams from one modality to another.

\textbf{MISA} \citep{hazarika2020misa} Modality-Invariant and -Specific Representations projects modalities to two distinct subspaces, modality-invariant and -specific subspace, to provide a holistic view of multimodal data.

\textbf{MAG-BERT} \citep{rahman2020integrating} Multimodal Adaption Gate for BERT applies multimodal adaptation gate to generate a shift for the internal representation of pre-trained Transformer models.

\textbf{Self-MM} \citep{yu2021learning} Self-Supervised Multi-Task Learning automatically generates uni-modal labels which are weight-adjusted by multimodal labels to learn consistency and difference across modalities.

\textbf{MMIM} \citep{han2021improving} Multimodal Mutual Information Maximization maintain task-related information through maximizing the mutual information in uni-modal input pairs and between multimodal fusion output and uni-modal input. 

\subsection{Results and Ablation Study}\label{sec_result}
In accord with baselines, our proposed model is run five times under the same setting of hyper-parameters and present the average performance in Table \ref{table1}. Significantly, MMCL outperforms SOTA in all metrics whether on CMU-MOSI or on CMU-MOSEI. These outcomes preliminarily demonstrate the effectiveness of our method in MSA task.

\begin{table}[htbp]
    \small
	\centering
	\setlength\tabcolsep{1.5pt}
	\begin{tabular}{lccccc}
		\toprule[1.5pt]
		\quad Description & Acc7$\uparrow$  & Acc2$\uparrow$ & F1$\uparrow$ & MAE$\downarrow$ & Corr$\uparrow$ \\
		\midrule[1.5pt]
		(1) UMCC\quad\quad\quad &&&&& \\
		rp Transformer & 51.0 &  81.9/84.7  &  82.2/84.6  & 0.567  & 0.741 \\
		w/o $\mathcal L_{uni}$ & 52.4  &  82.6/85.5  &  82.9/85.3  &  0.554 & 0.752 \\
		(2) CMCP\quad\quad\quad &&&&& \\
		w/o ICL & 52.0  &  81.7/84.7  &  82.0/84.6  &  0.550 & 0.755 \\
		w/o $\mathcal L_{cross}$ &  53.0 &  83.8/85.2  &  83.7/84.9  & 0.543 & 0.761 \\
		w/o $\mathcal L_{align}$ &  52.9 &  83.9/85.0  &  83.8/84.6  & 0.554 & 0.756 \\
		w/o $\mathcal L_{uniform}$ & 52.3 &  84.4/85.5  &  84.4/85.2  & 0.553 & 0.757 \\
		w/o SCL & 52.2 &  82.2/84.9  &  82.6/84.8  & 0.546 & 0.756 \\
		(3) No Contrast\quad & 51.3 &  81.2/84.3  &  81.6/84.0  & 0.564 & 0.746 \\
		\midrule[0.5pt]
		\quad\quad  MMCL & \textbf{53.6}  &  \textbf{84.8}/\textbf{85.9}  &  \textbf{84.8}/\textbf{85.7}  & \textbf{0.537}  &  \textbf{0.765}  \\
		\bottomrule[1.5pt]     
	\end{tabular}
	\caption{Ablation study of MMCL on CMU-MOSEI dataset. Note that "rp Transformer" denotes to replace uni-modal Transformers with linear layers. }
	\label{table2}
\end{table}

To further explore the contributions of proposed uni-modal coding and cross-modal predictive network with corresponding contrastive loss functions in MMCL, we carry out a series of ablation experiments on CMU-MOSEI. As shown in Table \ref{table2}, for UMCC, we firstly substitute uni-modal Transformers with linear layers and discover the great performance degradation of the proposed model, which means that the powerful Transformer encoders play a crucial part in uni-modal coding. Then we remove the uni-modal instance contrastive loss $\mathcal L_{uni}$ and note that although Acc2 (non-0) and F1 (non-0) drop slightly, the model have a sharp decline on other metrics. We consider that $\mathcal L_{uni}$ mainly focus on the inherent noise contained in uni-modal data. Without feature augmentation with $\mathcal L_{uni}$, the model is easier to be interrupted by hard samples such as the nearly neutral samples with sentiment labels around 0. For CMCP, we eliminate the loss terms in ICL individually and the results indicates that all the loss functions are useful for the sentiment prediction, which can be further observed in Section \ref{sec:loss}. Moreover, ablating either ICL or SCL hurt the model performance significantly. This demonstrates the effectiveness of two designed contrastive learning tasks to the cross-modal predictive network. Finally, we spot a clear drop in all metrics when taking off all contrastive losses and just training the model with just regression task loss $\mathcal L_{reg}$. The results of ablation study imply that each module and loss function are necessary and essential for MMCL to achieve the best performance.
\section{Further Analysis}
\begin{table*}[htbp]
	\centering
	\setlength\tabcolsep{2.5pt}
	\begin{tabular}{ccccccccccc}
		\toprule[1.5pt]
		\multirow{2}{*}{\makecell[c]{Unaligned\\ SOTA Models}} & \multicolumn{5}{c}{CMU-MOSI} & \multicolumn{5}{c}{CMU-MOSEI}\\
		\cmidrule(r){2-6}\cmidrule(r){7-11}
		& Acc7$\uparrow$  & Acc2$\uparrow$ & F1$\uparrow$ & MAE$\downarrow$ & Corr$\uparrow$ & Acc7$\uparrow$  & Acc2$\uparrow$ & F1$\uparrow$ & MAE$\downarrow$ & Corr$\uparrow$\\
		\midrule[1.5pt]
		MulT &  36.5 &  78.8/80.2  &  78.7/80.2  & 0.899  & 0.687  &
		52.5 &  81.0/84.0  &  81.4/84.0  & 0.561  & 0.737 \\
		MISA &  43.6 &  81.6/83.7  &  81.5/83.6  & 0.752  & 0.779 &  
		52.0 &  81.1/84.3  &  81.6/84.4  & 0.554 & 0.753  \\
		Self-MM &  45.9 &  82.6/84.9  &  82.5/84.8  &  0.722 & 0.789  & 
		53.1 &  82.2/84.9  &  82.4/85.0  & 0.540 & 0.760 \\
		MMIM &  45.5 &  82.6/85.0  &  82.5/84.9  & 0.726  & 0.786  & 
		53.2 &  82.2/85.1  &  82.6/85.0  & 0.544  & 0.756 \\
		\midrule[0.5pt]
		MMCL (ours) & \textbf{46.8}  &  \textbf{83.8}/\textbf{85.8}  &  \textbf{83.7}/\textbf{85.7}  &  \textbf{0.692} & \textbf{0.800}  & \textbf{53.4}  &  \textbf{84.5}/\textbf{85.7}  &  \textbf{84.6}/\textbf{85.6}  & \textbf{0.538}  &  \textbf{0.765}\\
		\bottomrule[1.5pt]     
	\end{tabular}
	\caption{Performance Comparison between MMCL and baselines on CMU-MOSI and CMU-MOSEI datasets in a token-unaligned manner with token-unaligned multimodal data.}
	\label{table3}
\end{table*}
\subsection{Losses Tracing}\label{sec:loss}
\begin{figure}[htbp]
	\centering 
	\includegraphics[scale=0.26]{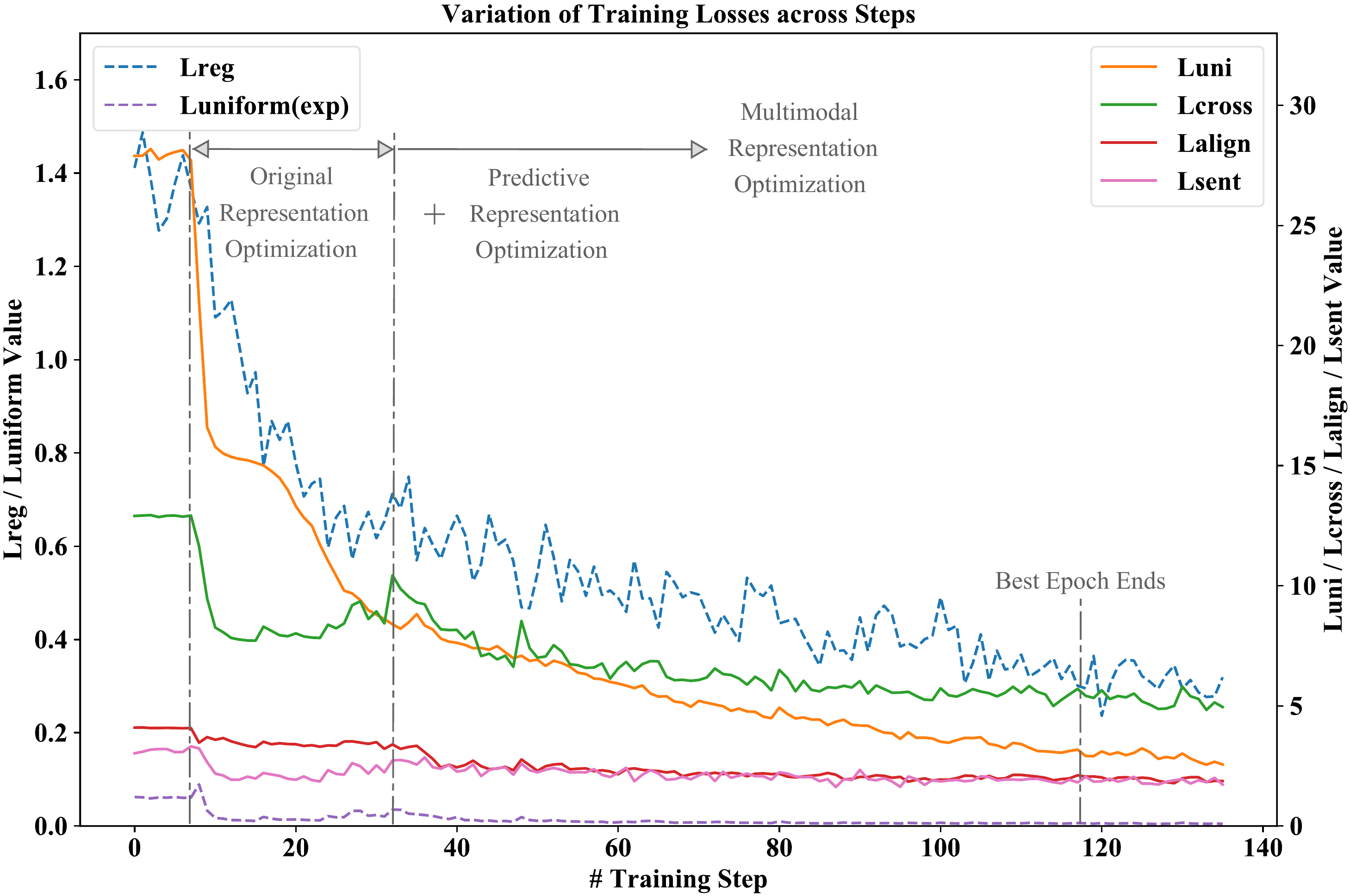}
	\caption{Visualization of losses changing as training proceeds on CMU-MOSI. The values for plotting are the average losses in a constant interval of every 5 steps and $\mathcal L_{uniform}$ is exponentiated for plotting purposes.}
	\label{vis_loss}
\end{figure}
To better understand how every contrastive loss works, we visualize the variation of all losses during training in Figure \ref{vis_loss}. At the beginning of training, MMCL tends to optimize the original representation for each modality since all losses decrease almost concurrently. Then when other losses flatten out, $\mathcal L_{uni}$ continue to decline which means UMCC starts to filter out the inherent noises of hard samples. Next, we observe that $\mathcal L_{cross}$ and  $\mathcal L_{uniform}$ increased slightly where MMCL attempt to optimize the predictive representation, which is due to the extreme difficulty for the predictive network to inference across different modalities. Succeeding in reducing the modality gap and promoting the network to converge, the designed contrastive tasks ICL and SCL make the losses drop again with a smaller slope. Finally, the training of MMCL ends at the best epochs when the task loss on the validation set reaches the minimum. In the optimization process of multimodal representation, the designed losses are productive to the model in an alternating manner.
\subsection{Representation Visualization}
\begin{figure} 
    \centering    
    \subfigure[MMCL without CL] {
     \label{vis_feature_wo_CL}
    \includegraphics[width=0.46\columnwidth]{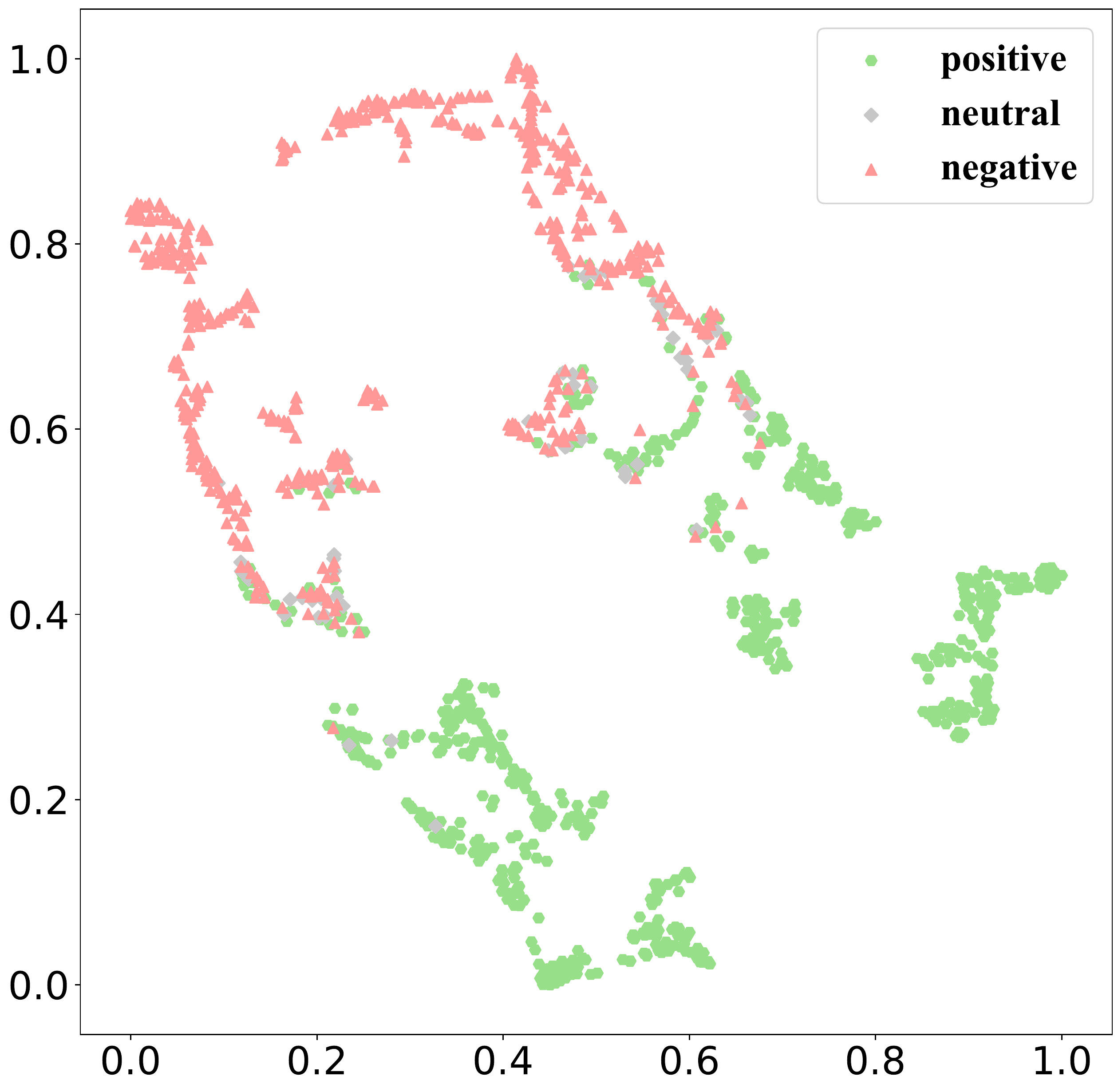}  
    }
    \subfigure[MMCL with CL] { 
    \label{vis_feature_w_CL}
    \includegraphics[width=0.46\columnwidth]{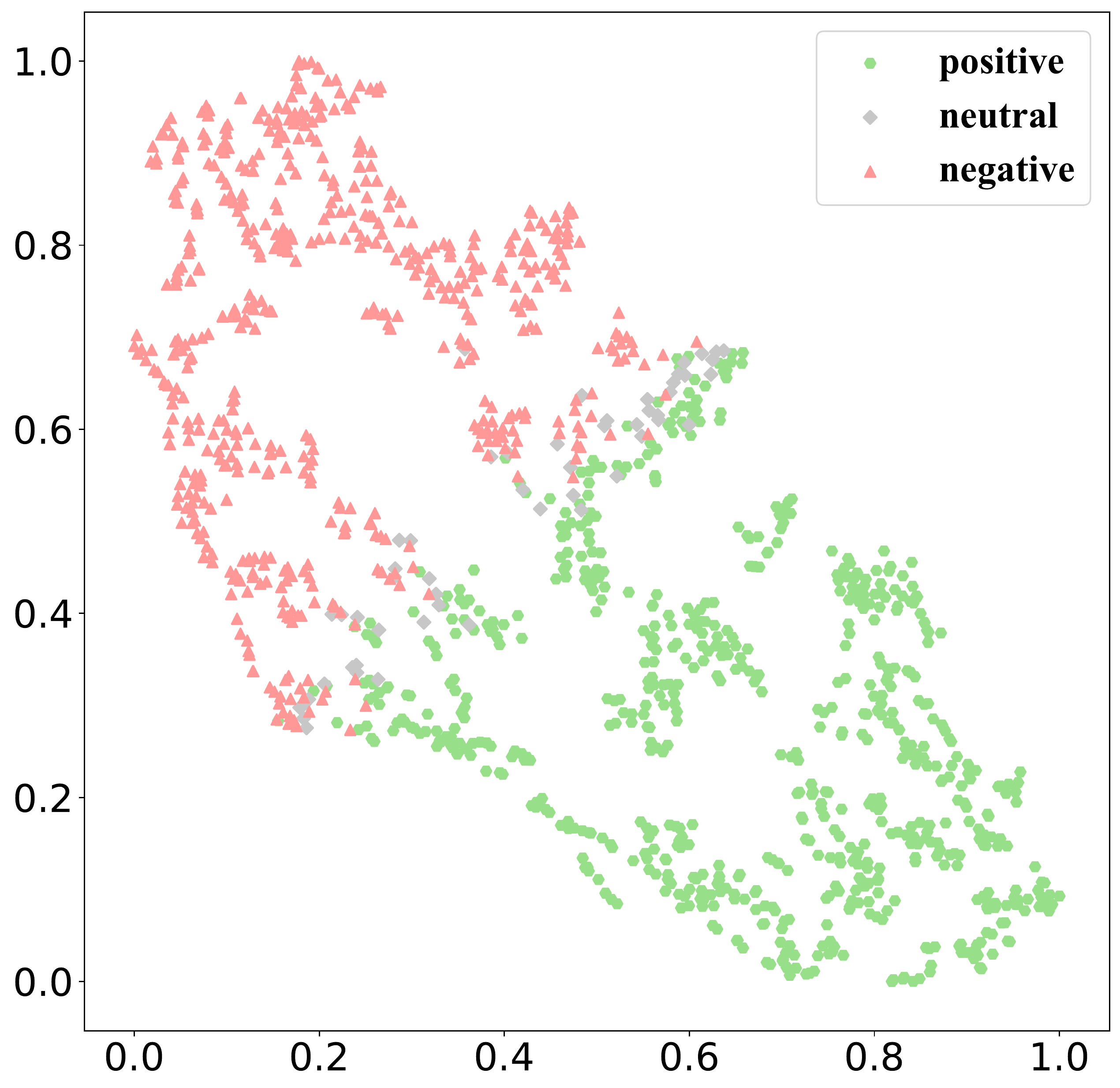}     
    }
    \caption{T-SNE \citep{hinton2008visualizing} visualization of multimodal representation in the embedding space on the training set of CMU-MOSI.}
    \label{vis_feature}
\end{figure}

Figure \ref{vis_feature} displays the visualization of fusion multimodal representation $F_M$ calculated by MMCL with contrastive learning losses or not. Without contrastive learning, the representations in the same batch are clustered too tight while far away from each other in the different batches, which means that the model is inclined to overfit the training dataset. Meanwhile, the hard samples with nearly neutral sentiment are indistinguishable for the model due to the absence of $\mathcal L_{uni}$ and $\mathcal L_{sent}$. After introducing designed contrastive learning, MMCL is more capable of mining the commonalities between different instances with the same sentiment polarity, while retaining the specific features contained in different instances. In the interim, the hard samples are more distinguishable for corresponding sentiment class which also proves the effectiveness of the designed contrastive learning tasks on representation learning.
\subsection{Token-Unaligned Exploration}
Additionally, we explore the performance of MMCL framework on token-unaligned data as shown in Table \ref{table3}. All models are run five times the same as the experiments on token-aligned data. Basically, the performance of models trained on token-unaligned data are similar to the ones trained on token-aligned data with a slight drop on some metrics. Nevertheless, compared with unaligned state-of-the-art models, MMCL still achieves the best result on all metrics, which manifests the significant power of the proposed framework. 
\section{Conclusion}
In this paper, we present a novel framework named MultiModal Contrastive Learning (MMCL) which efficiently adopts contrastive learning in multimodal sentiment analysis. MMCL consists of Uni-Modal Contrastive Coding (UMCC) which learns robust uni-modal representations by focusing on informative features and reducing the interference of inherent noise for acoustic and visual modalities, and Cross-Modal Contrastive Prediction (CMCP) which construct a powerful pseudo siamese predictive network to learn commonalities and interactive dynamics across different modalities. Besides, we propose two effective contrastive learning tasks, instance- and sentiment-based contrastive learning, to help the predictive network maintain modality-specific dynamics and learn sentiment-related information in multimodal data, which improve the performance of the model significantly. The extensive experiments conducted on CMU-MOSI and CMU-MOSEI demonstrate the superiority of our model and the efficacy of contrastive learning. Moreover, the further visualization of losses and representation space provide comprehensive insight into our model. We believe our work can boost the creativity in contrastive learning and multimodal representation learning in the future.
\section*{Limitations}
Below we describe the limitations of the proposed model in our view and suggest directions for future work. Firstly, there are still a lot of data augmentation strategy can be further examined in the process of encoding uni-modal representation. Then, with the proposed powerful predictive network, we can extend the predictive method into dealing with missing modality problem which is a more challenging research emphasis. At last, the training process of proposed model is still a supervised leaning process. Since contrastive learning is increasingly adopted in semi- and self-supervised learning which burst out with great potential in computer vision, we attempt to remove the supervised labels and explore its capability in the field of multimodal in the future work.

\section*{Acknowledgements}
This work was supported in part by the National Natural Science Foundation of China under Grant 62076262.

\bibliography{anthology,custom}

\begin{thebibliography}{38}
\expandafter\ifx\csname natexlab\endcsname\relax\def\natexlab#1{#1}\fi

\bibitem[{Bromley et~al.(1993)Bromley, Guyon, LeCun, S{\"a}ckinger, and
  Shah}]{bromley1993signature}
Jane Bromley, Isabelle Guyon, Yann LeCun, Eduard S{\"a}ckinger, and Roopak
  Shah. 1993.
\newblock \href
  {https://proceedings.neurips.cc/paper/1993/file/288cc0ff022877bd3df94bc9360b9c5d-Paper.pdf}
  {Signature verification using a" siamese" time delay neural network}.
\newblock \emph{Advances in neural information processing systems}, 6.

\bibitem[{Cohn and Kumar(2007)}]{cohn2007universally}
Henry Cohn and Abhinav Kumar. 2007.
\newblock \href {https://arxiv.org/pdf/math/0607446.pdf} {Universally optimal
  distribution of points on spheres}.
\newblock \emph{Journal of the American Mathematical Society}, 20(1):99--148.

\bibitem[{Degottex et~al.(2014)Degottex, Kane, Drugman, Raitio, and
  Scherer}]{degottex2014covarep}
Gilles Degottex, John Kane, Thomas Drugman, Tuomo Raitio, and Stefan Scherer.
  2014.
\newblock \href {https://doi.org/10.1109/ICASSP.2014.6853739} {Covarep — a
  collaborative voice analysis repository for speech technologies}.
\newblock In \emph{2014 IEEE International Conference on Acoustics, Speech and
  Signal Processing (ICASSP)}, pages 960--964.

\bibitem[{Devlin et~al.(2019)Devlin, Chang, Lee, and
  Toutanova}]{devlin2019bert}
Jacob Devlin, Ming-Wei Chang, Kenton Lee, and Kristina Toutanova. 2019.
\newblock \href {https://doi.org/10.18653/v1/N19-1423} {{BERT}: Pre-training of
  deep bidirectional transformers for language understanding}.
\newblock In \emph{Proceedings of the 2019 Conference of the North {A}merican
  Chapter of the Association for Computational Linguistics: Human Language
  Technologies, Volume 1 (Long and Short Papers)}, pages 4171--4186,
  Minneapolis, Minnesota. Association for Computational Linguistics.

\bibitem[{Gao et~al.(2021)Gao, Yao, and Chen}]{gao2021simcse}
Tianyu Gao, Xingcheng Yao, and Danqi Chen. 2021.
\newblock \href {https://doi.org/10.18653/v1/2021.emnlp-main.552} {{S}im{CSE}:
  Simple contrastive learning of sentence embeddings}.
\newblock In \emph{Proceedings of the 2021 Conference on Empirical Methods in
  Natural Language Processing}, pages 6894--6910, Online and Punta Cana,
  Dominican Republic. Association for Computational Linguistics.

\bibitem[{Hadsell et~al.(2006)Hadsell, Chopra, and
  LeCun}]{hadsell2006dimensionality}
R.~Hadsell, S.~Chopra, and Y.~LeCun. 2006.
\newblock \href {https://doi.org/10.1109/CVPR.2006.100} {Dimensionality
  reduction by learning an invariant mapping}.
\newblock In \emph{2006 IEEE Computer Society Conference on Computer Vision and
  Pattern Recognition (CVPR'06)}, volume~2, pages 1735--1742.

\bibitem[{Han et~al.(2021)Han, Chen, and Poria}]{han2021improving}
Wei Han, Hui Chen, and Soujanya Poria. 2021.
\newblock \href {https://doi.org/10.18653/v1/2021.emnlp-main.723} {Improving
  multimodal fusion with hierarchical mutual information maximization for
  multimodal sentiment analysis}.
\newblock In \emph{Proceedings of the 2021 Conference on Empirical Methods in
  Natural Language Processing}, pages 9180--9192, Online and Punta Cana,
  Dominican Republic. Association for Computational Linguistics.

\bibitem[{Hazarika et~al.(2020)Hazarika, Zimmermann, and
  Poria}]{hazarika2020misa}
Devamanyu Hazarika, Roger Zimmermann, and Soujanya Poria. 2020.
\newblock \href {https://doi.org/10.1145/3394171.3413678} {Misa:
  Modality-invariant and-specific representations for multimodal sentiment
  analysis}.
\newblock In \emph{Proceedings of the 28th ACM International Conference on
  Multimedia}, pages 1122--1131.

\bibitem[{He et~al.(2020)He, Fan, Wu, Xie, and Girshick}]{he2020momentum}
Kaiming He, Haoqi Fan, Yuxin Wu, Saining Xie, and Ross Girshick. 2020.
\newblock \href {https://doi.org/10.1109/CVPR42600.2020.00975} {Momentum
  contrast for unsupervised visual representation learning}.
\newblock In \emph{2020 IEEE/CVF Conference on Computer Vision and Pattern
  Recognition (CVPR)}, pages 9726--9735.

\bibitem[{Hinton et~al.(2015)Hinton, Vinyals, and Dean}]{hinton2015distilling}
Geoffrey Hinton, Oriol Vinyals, and Jeff Dean. 2015.
\newblock \href {https://arxiv.org/pdf/1503.02531} {Distilling the knowledge in
  a neural network}.
\newblock \emph{arXiv preprint arXiv:1503.02531}, 2.

\bibitem[{Hochreiter and Schmidhuber(1997)}]{hochreiter1997long}
Sepp Hochreiter and Jürgen Schmidhuber. 1997.
\newblock \href {https://doi.org/10.1162/neco.1997.9.8.1735} {Long short-term
  memory}.
\newblock \emph{Neural Computation}, 9(8):1735--1780.

\bibitem[{iMotions 2017()}]{iMotions}
iMotions 2017.
\newblock Facial expression analysis.
\newblock [Online].
\newblock \url{https://imotions.com/}.

\bibitem[{Khosla et~al.(2020)Khosla, Teterwak, Wang, Sarna, Tian, Isola,
  Maschinot, Liu, and Krishnan}]{khosla2020supervised}
Prannay Khosla, Piotr Teterwak, Chen Wang, Aaron Sarna, Yonglong Tian, Phillip
  Isola, Aaron Maschinot, Ce~Liu, and Dilip Krishnan. 2020.
\newblock \href
  {https://proceedings.neurips.cc/paper/2020/file/d89a66c7c80a29b1bdbab0f2a1a94af8-Paper.pdf}
  {Supervised contrastive learning}.
\newblock In \emph{Advances in Neural Information Processing Systems},
  volume~33, pages 18661--18673. Curran Associates, Inc.

\bibitem[{Kingma and Ba(2014)}]{kingma2014adam}
Diederik~P Kingma and Jimmy Ba. 2014.
\newblock \href {https://arxiv.org/abs/1412.6980} {Adam: A method for
  stochastic optimization}.
\newblock \emph{arXiv preprint arXiv:1412.6980}.

\bibitem[{Liu et~al.(2018)Liu, Shen, Lakshminarasimhan, Liang, Bagher~Zadeh,
  and Morency}]{liu2018efficient}
Zhun Liu, Ying Shen, Varun~Bharadhwaj Lakshminarasimhan, Paul~Pu Liang, AmirAli
  Bagher~Zadeh, and Louis-Philippe Morency. 2018.
\newblock \href {https://doi.org/10.18653/v1/P18-1209} {Efficient low-rank
  multimodal fusion with modality-specific factors}.
\newblock In \emph{Proceedings of the 56th Annual Meeting of the Association
  for Computational Linguistics (Volume 1: Long Papers)}, pages 2247--2256,
  Melbourne, Australia. Association for Computational Linguistics.

\bibitem[{Morency et~al.(2011)Morency, Mihalcea, and
  Doshi}]{morency2011towards}
Louis-Philippe Morency, Rada Mihalcea, and Payal Doshi. 2011.
\newblock \href {https://doi.org/10.1145/2070481.2070509} {Towards multimodal
  sentiment analysis: Harvesting opinions from the web}.
\newblock In \emph{Proceedings of the 13th International Conference on
  Multimodal Interfaces}, ICMI '11, page 169–176, New York, NY, USA.
  Association for Computing Machinery.

\bibitem[{Oord et~al.(2018)Oord, Li, and Vinyals}]{van2018representation}
Aaron van~den Oord, Yazhe Li, and Oriol Vinyals. 2018.
\newblock \href {https://arxiv.org/pdf/1807.03748.pdf} {Representation learning
  with contrastive predictive coding}.
\newblock \emph{arXiv e-prints}, pages arXiv--1807.

\bibitem[{Poria et~al.(2017)Poria, Cambria, Hazarika, Majumder, Zadeh, and
  Morency}]{poria2017context}
Soujanya Poria, Erik Cambria, Devamanyu Hazarika, Navonil Majumder, Amir Zadeh,
  and Louis-Philippe Morency. 2017.
\newblock \href {https://doi.org/10.18653/v1/P17-1081} {Context-dependent
  sentiment analysis in user-generated videos}.
\newblock In \emph{Proceedings of the 55th Annual Meeting of the Association
  for Computational Linguistics (Volume 1: Long Papers)}, pages 873--883,
  Vancouver, Canada. Association for Computational Linguistics.

\bibitem[{Poria et~al.(2020)Poria, Hazarika, Majumder, and
  Mihalcea}]{poria2020beneath}
Soujanya Poria, Devamanyu Hazarika, Navonil Majumder, and Rada Mihalcea. 2020.
\newblock \href {https://doi.org/10.1109/TAFFC.2020.3038167} {Beneath the tip
  of the iceberg: Current challenges and new directions in sentiment analysis
  research}.
\newblock \emph{IEEE Transactions on Affective Computing}.

\bibitem[{Radford et~al.(2021)Radford, Kim, Hallacy, Ramesh, Goh, Agarwal,
  Sastry, Askell, Mishkin, Clark, Krueger, and Sutskever}]{radford2021learning}
Alec Radford, Jong~Wook Kim, Chris Hallacy, Aditya Ramesh, Gabriel Goh,
  Sandhini Agarwal, Girish Sastry, Amanda Askell, Pamela Mishkin, Jack Clark,
  Gretchen Krueger, and Ilya Sutskever. 2021.
\newblock \href {https://proceedings.mlr.press/v139/radford21a.html} {Learning
  transferable visual models from natural language supervision}.
\newblock In \emph{Proceedings of the 38th International Conference on Machine
  Learning}, volume 139 of \emph{Proceedings of Machine Learning Research},
  pages 8748--8763. PMLR.

\bibitem[{Rahman et~al.(2020)Rahman, Hasan, Lee, Bagher~Zadeh, Mao, Morency,
  and Hoque}]{rahman2020integrating}
Wasifur Rahman, Md~Kamrul Hasan, Sangwu Lee, AmirAli Bagher~Zadeh, Chengfeng
  Mao, Louis-Philippe Morency, and Ehsan Hoque. 2020.
\newblock \href {https://doi.org/10.18653/v1/2020.acl-main.214} {Integrating
  multimodal information in large pretrained transformers}.
\newblock In \emph{Proceedings of the 58th Annual Meeting of the Association
  for Computational Linguistics}, pages 2359--2369, Online. Association for
  Computational Linguistics.

\bibitem[{Shen et~al.(2020)Shen, Zheng, Shen, Qu, and Chen}]{shen2020simple}
Dinghan Shen, Mingzhi Zheng, Yelong Shen, Yanru Qu, and Weizhu Chen. 2020.
\newblock \href {https://arxiv.org/pdf/2009.13818.pdf} {A simple but
  tough-to-beat data augmentation approach for natural language understanding
  and generation}.
\newblock \emph{arXiv preprint arXiv:2009.13818}.

\bibitem[{Tsai et~al.(2019{\natexlab{a}})Tsai, Bai, Liang, Kolter, Morency, and
  Salakhutdinov}]{tsai2019multimodal}
Yao-Hung~Hubert Tsai, Shaojie Bai, Paul~Pu Liang, J~Zico Kolter, Louis-Philippe
  Morency, and Ruslan Salakhutdinov. 2019{\natexlab{a}}.
\newblock \href {https://doi.org/10.18653/v1/P19-1656} {Multimodal transformer
  for unaligned multimodal language sequences}.
\newblock In \emph{Proceedings of the 57th Annual Meeting of the Association
  for Computational Linguistics}, pages 6558--6569, Florence, Italy.
  Association for Computational Linguistics.

\bibitem[{Tsai et~al.(2019{\natexlab{b}})Tsai, Liang, Zadeh, Morency, and
  Salakhutdinov}]{tsai2019learning}
Yao-Hung~Hubert Tsai, Paul~Pu Liang, Amir Zadeh, Louis-Philippe Morency, and
  Ruslan Salakhutdinov. 2019{\natexlab{b}}.
\newblock \href {https://openreview.net/forum?id=rygqqsA9KX} {Learning
  factorized multimodal representations}.
\newblock In \emph{International Conference on Learning Representations}.

\bibitem[{Tsai et~al.(2020)Tsai, Ma, Yang, Salakhutdinov, and
  Morency}]{tsai2020multimodal}
Yao-Hung~Hubert Tsai, Martin Ma, Muqiao Yang, Ruslan Salakhutdinov, and
  Louis-Philippe Morency. 2020.
\newblock \href {https://doi.org/10.18653/v1/2020.emnlp-main.143} {Multimodal
  routing: Improving local and global interpretability of multimodal language
  analysis}.
\newblock In \emph{Proceedings of the 2020 Conference on Empirical Methods in
  Natural Language Processing (EMNLP)}, pages 1823--1833, Online. Association
  for Computational Linguistics.

\bibitem[{van~der Maaten and Hinton(2008)}]{hinton2008visualizing}
Laurens van~der Maaten and Geoffrey Hinton. 2008.
\newblock \href {http://jmlr.org/papers/v9/vandermaaten08a.html} {Visualizing
  data using t-sne}.
\newblock \emph{Journal of Machine Learning Research}, 9(86):2579--2605.

\bibitem[{Wang and Isola(2020)}]{wang2020understanding}
Tongzhou Wang and Phillip Isola. 2020.
\newblock \href {http://proceedings.mlr.press/v119/wang20k/wang20k.pdf}
  {Understanding contrastive representation learning through alignment and
  uniformity on the hypersphere}.
\newblock In \emph{International Conference on Machine Learning}, pages
  9929--9939. PMLR.

\bibitem[{Wu et~al.(2022)Wu, Seetharaman, Kumar, and Bello}]{wu2022wav2clip}
Ho-Hsiang Wu, Prem Seetharaman, Kundan Kumar, and Juan~Pablo Bello. 2022.
\newblock \href {https://doi.org/10.1109/ICASSP43922.2022.9747669} {Wav2clip:
  Learning robust audio representations from clip}.
\newblock In \emph{ICASSP 2022-2022 IEEE International Conference on Acoustics,
  Speech and Signal Processing (ICASSP)}, pages 4563--4567. IEEE.

\bibitem[{Wu et~al.(2021)Wu, Lin, Zhao, Qin, and Zhu}]{wu2021text}
Yang Wu, Zijie Lin, Yanyan Zhao, Bing Qin, and Li-Nan Zhu. 2021.
\newblock \href {https://doi.org/10.18653/v1/2021.findings-acl.417} {A
  text-centered shared-private framework via cross-modal prediction for
  multimodal sentiment analysis}.
\newblock In \emph{Findings of the Association for Computational Linguistics:
  ACL-IJCNLP 2021}, pages 4730--4738, Online. Association for Computational
  Linguistics.

\bibitem[{Wu et~al.(2018)Wu, Xiong, Yu, and Lin}]{wu2018unsupervised}
Zhirong Wu, Yuanjun Xiong, Stella~X Yu, and Dahua Lin. 2018.
\newblock \href
  {https://openaccess.thecvf.com/content_cvpr_2018/papers/Wu_Unsupervised_Feature_Learning_CVPR_2018_paper.pdf}
  {Unsupervised feature learning via non-parametric instance discrimination}.
\newblock In \emph{Proceedings of the IEEE conference on computer vision and
  pattern recognition}, pages 3733--3742.

\bibitem[{Ye et~al.(2019)Ye, Zhang, Yuen, and Chang}]{ye2019unsupervised}
Mang Ye, Xu~Zhang, Pong~C Yuen, and Shih-Fu Chang. 2019.
\newblock \href
  {https://openaccess.thecvf.com/content_CVPR_2019/papers/Ye_Unsupervised_Embedding_Learning_via_Invariant_and_Spreading_Instance_Feature_CVPR_2019_paper.pdf}
  {Unsupervised embedding learning via invariant and spreading instance
  feature}.
\newblock In \emph{Proceedings of the IEEE/CVF Conference on Computer Vision
  and Pattern Recognition}, pages 6210--6219.

\bibitem[{Yu et~al.(2021)Yu, Xu, Yuan, and Wu}]{yu2021learning}
Wenmeng Yu, Hua Xu, Ziqi Yuan, and Jiele Wu. 2021.
\newblock \href {https://ojs.aaai.org/index.php/AAAI/article/view/17289/17096}
  {Learning modality-specific representations with self-supervised multi-task
  learning for multimodal sentiment analysis}.
\newblock In \emph{Proceedings of the AAAI Conference on Artificial
  Intelligence}, volume~35, pages 10790--10797.

\bibitem[{Zadeh et~al.(2017)Zadeh, Chen, Poria, Cambria, and
  Morency}]{zadeh2017tensor}
Amir Zadeh, Minghai Chen, Soujanya Poria, Erik Cambria, and Louis-Philippe
  Morency. 2017.
\newblock \href {https://doi.org/10.18653/v1/D17-1115} {Tensor fusion network
  for multimodal sentiment analysis}.
\newblock In \emph{Proceedings of the 2017 Conference on Empirical Methods in
  Natural Language Processing}, pages 1103--1114, Copenhagen, Denmark.
  Association for Computational Linguistics.

\bibitem[{Zadeh et~al.(2018{\natexlab{a}})Zadeh, Liang, Mazumder, Poria,
  Cambria, and Morency}]{zadeh2018memory}
Amir Zadeh, Paul~Pu Liang, Navonil Mazumder, Soujanya Poria, Erik Cambria, and
  Louis-Philippe Morency. 2018{\natexlab{a}}.
\newblock \href {https://ojs.aaai.org/index.php/AAAI/article/view/12021/11880}
  {Memory fusion network for multi-view sequential learning}.
\newblock In \emph{Proceedings of the AAAI Conference on Artificial
  Intelligence}, volume~32.

\bibitem[{Zadeh et~al.(2018{\natexlab{b}})Zadeh, Liang, Poria, Cambria, and
  Morency}]{zadeh2018multimodal}
Amir Zadeh, Paul~Pu Liang, Soujanya Poria, Erik Cambria, and Louis-Philippe
  Morency. 2018{\natexlab{b}}.
\newblock \href {https://doi.org/10.18653/v1/P18-1208} {Multimodal language
  analysis in the wild: {CMU}-{MOSEI} dataset and interpretable dynamic fusion
  graph}.
\newblock In \emph{Proceedings of the 56th Annual Meeting of the Association
  for Computational Linguistics (Volume 1: Long Papers)}, pages 2236--2246,
  Melbourne, Australia. Association for Computational Linguistics.

\bibitem[{Zadeh et~al.(2018{\natexlab{c}})Zadeh, Liang, Poria, Vij, Cambria,
  and Morency}]{zadeh2018multi}
Amir Zadeh, Paul~Pu Liang, Soujanya Poria, Prateek Vij, Erik Cambria, and
  Louis-Philippe Morency. 2018{\natexlab{c}}.
\newblock \href {https://doi.org/10.1609/aaai.v32i1.12024} {Multi-attention
  recurrent network for human communication comprehension}.
\newblock In \emph{Proceedings of the AAAI Conference on Artificial
  Intelligence}, volume~32.

\bibitem[{Zadeh et~al.(2016)Zadeh, Zellers, Pincus, and
  Morency}]{zadeh2016mosi}
Amir Zadeh, Rowan Zellers, Eli Pincus, and Louis-Philippe Morency. 2016.
\newblock \href {https://arxiv.org/ftp/arxiv/papers/1606/1606.06259.pdf} {Mosi:
  multimodal corpus of sentiment intensity and subjectivity analysis in online
  opinion videos}.
\newblock \emph{arXiv preprint arXiv:1606.06259}.

\bibitem[{Zhao et~al.(2021)Zhao, Li, and Jin}]{zhao2021missing}
Jinming Zhao, Ruichen Li, and Qin Jin. 2021.
\newblock \href {https://doi.org/10.18653/v1/2021.acl-long.203} {Missing
  modality imagination network for emotion recognition with uncertain missing
  modalities}.
\newblock In \emph{Proceedings of the 59th Annual Meeting of the Association
  for Computational Linguistics and the 11th International Joint Conference on
  Natural Language Processing (Volume 1: Long Papers)}, pages 2608--2618,
  Online. Association for Computational Linguistics.

\end{thebibliography}
\bibliographystyle{acl_natbib}

\appendix

\section{Appendix}\label{sec:appendix}
\subsection{Implementation Details}
To fairly compare the effectiveness of baselines and our proposed frameworks, we reproduce the baselines based on the same pre-trained BERT language model as text features encoder and ran hyper-parameter grid search for the best results. All models are trained on a single GTX 1080Ti GPU. As shown in Table \ref{table1}, noted that except for unaligned models TFN and LMF, the multimodal data are token-aligned in the pre-processing stage.

We perform grid-search over finite sets of options for hyper-parameters, including $\tau$ in $\{0.05, 0.1, 0.7\}$, $\mu$ in $\{0.6, 0.7, 0.8\}$, $\eta$ in $\{0.8, 0.9, 1.0\}$, $\alpha$ in $\{0.8, 0.9, 1.0\}$, $\beta$ in $\{0.6, 0.75, 0.9\}$ and $\gamma$ in $\{0.05, 0.1, 0.2\}$ and dropout values in $\{0.1, 0.6\}$, to select the best model. For optimization, we apply Adam\citep{kingma2014adam} as the optimizer with a linear warmup learning rate scheduler, using the highest learning rate 5e-5 for BERT finetuning and 5e-3 on CMU-MOSI and 1e-3 on CMU-MOSEI for other parameters. The batch size on both datasets is set as 32. In UMCC, the uni-modal Transformers have a stack of 3 identical layers with 8 parallel attention heads for both acoustic and visual modalities. The random cutoff ratio is set as 0.2. In CMCP, the pseudo siamese CNNs have five convolutional layers with strides [5, 4, 2, 2, 2], filter-sizes [10, 8, 4, 4, 4] and 256 hidden units with $Tanh$ activation layers while the sLSTM as auto-regressive model has 128 dimensional hidden states.
\subsection{Explanation of Eq.\ref{equ6}-\ref{equ7}}
Eq.\ref{equ6} aims at aligning the query-positive samples pairs in the contrastive learning due to the fact that the feature distance of samples pairs should be minimized if they are from query-positive samples pairs, which are the prediction and target representations in CMCP. Here $\lambda\in\{1, 2\}$ illustrates the form of loss function as L1 and L2 loss, respectively. 

Eq. \ref{equ7} is proposed according to the thought that the feature distribution of the predictive and target representations need to preserve maximal modality-specific information, which is the uniform distribution on the unit hypersphere. For this purpose, the Gaussian potential kernel \citep{cohn2007universally}, known as the Radial Basis Function kernel $K$ are considered as:
\begin{equation}
	K \triangleq e^{-\kappa\Vert P-G\Vert_2^2},\ \ \kappa > 0
\end{equation}
Then in Eq. \ref{equ7}, the uniformity is further defined as the logarithm of the average pairwise Gaussian potential, which is properly tied with the uniform distribution on the unit hypersphere.

\subsection{Ablation Study on Multiple Modalities}
\begin{table}[h]
	\centering
	\setlength\tabcolsep{1.5pt}
	\begin{tabular}{cccccc}
		\toprule[1.5pt]
		\fontsize{9.5pt}{\baselineskip}\selectfont{Modality} & Acc7$\uparrow$  & Acc2$\uparrow$ & F1$\uparrow$ & MAE$\downarrow$ & Corr$\uparrow$\\
		\midrule[1.5pt]
		\fontsize{9.5pt}{\baselineskip}\selectfont{$A$} & 41.6 & 70.0/62.8 & 60.5/51.1 & 0.835 & 0.116 \\
		\fontsize{9.5pt}{\baselineskip}\selectfont{$V$} & 40.9 & 70.9/62.8 & 59.2/48.9 & 0.841 & 0.109 \\
		\fontsize{9.5pt}{\baselineskip}\selectfont{$T$} & 52.6 & 80.7/85.3 & 81.4/85.4 & 0.552 & 0.761 \\
		\fontsize{9.5pt}{\baselineskip}\selectfont{$A+V$} & 41.4 & 71.0/62.9 & 59.0/48.5 & 0.837 & 0.148 \\
		\fontsize{9.5pt}{\baselineskip}\selectfont{$T+A$} & 53.3 & 82.4/85.5 & 82.8/85.4 & 0.540 & 0.762 \\
		\fontsize{9.5pt}{\baselineskip}\selectfont{$T+V$} & \textbf{53.6} & 83.6/\textbf{85.9} & 83.9/\textbf{85.9} & 0.538 & 0.763 \\
		\fontsize{8pt}{\baselineskip}\selectfont{$T+A+V$} & \textbf{53.6}  &  \textbf{84.8}/\textbf{85.9}  &  \textbf{84.8}/85.7  & \textbf{0.537}  &  \textbf{0.765}\\
		\bottomrule[1.5pt]     
	\end{tabular}
	\caption{Ablation study of MMCL on multiple modalities on CMU-MOSEI dataset. Note that $T,A,V$ represent textual, acoustic and vision modality, respectively.}
	\label{table4}
\end{table}
To evaluate the contribution of each modality, we conduct ablation study of MMCL on multiple modalities as shown in Table \ref{table4}. For better observing the effect in performance without changing the architecture of the proposed model, we remove modality by setting the corresponding representation to all zero vector. Firstly, we conclude that the multimodal representation learned by MMCL provides the best performance, indicating that the designed network and multimodal contrastive learning tasks can effectively capture complementary features across different modalities. In addition, the performance dramatically drops when the textual modality is removed which do not happen when removing the other modalities. The results show that the textual modality palys a significant dominant part in MSA task due to inherently better data quality and powerful pre-trained model. Although UMCC can reduce the interference of uni-modal noise and focus more on the task-related features for acoustic and visual modalities, the randomly initialized uni-modal feature extractor still lack of comparable information extraction capability with BERT.

\subsection{Discussion on Varying Hyperparameters}
As shown in Table \ref{table5}, to further explore the relationship of different losses, we conduct ablation experiments of MMCL on CMU-MOSEI dataset under different loss hyperparameters setting. We can observe that there are no significant performance drop in Table \ref{table5}, which concludes that the performance of MMCL is independent from the balance hyperparameters setting. Due to the different losses contribute to different modules of MMCL, the variation of hyperparameters can even help the model achieve better accuracy in several metrics. 
\begin{table}[h]
	\centering
	\setlength\tabcolsep{1.5pt}
	\begin{tabular}{cccccc}
		\toprule[1.5pt]
		\multicolumn{6}{c}{Varying Hyperparameters} \\
		\midrule[1.5pt] 
		\fontsize{9.5pt}{\baselineskip}\selectfont{Value} & Acc7$\uparrow$  & Acc2$\uparrow$ & F1$\uparrow$ & MAE$\downarrow$ & Corr$\uparrow$\\
		\midrule[1.5pt]
		$\mu=0.4$ & \textbf{54.1} & 83.2/85.5 & 83.5/85.4 & 0.536 & \textbf{0.768} \\
		$\mu=0.5$ & 53.5 & 83.1/\textbf{86.1} & 83.5/\textbf{86.0} & 0.537 & 0.760 \\
		$\mu=0.6$ & 53.6 & 83.6/85.9 & 83.8/85.7 & 0.536 & 0.763 \\
		$\mu=0.7$ & 53.6 & \textbf{84.8}/85.9 & \textbf{84.8}/85.6 & 0.537 & 0.765 \\
		$\mu=0.8$ & 54.1 & 83.0/85.6 & 83.3/85.5 & \textbf{0.533} & 0.764 \\
		$\mu=0.9$ & 52.7 & 83.8/85.6 & 84.0/85.5 & 0.543 & 0.764 \\
		\midrule[1.5pt]
		$\eta=0.4$ & 53.7 & 83.0/85.7 & 83.3/85.6 & 0.532 & 0.765 \\
		$\eta=0.6$ & 52.4 & 84.5/84.9 & 84.4/84.6 & 0.544 & 0.760 \\
		$\eta=0.8$ & \textbf{54.1} & 84.1/85.1 & 84.2/84.9 & \textbf{0.536} & 0.763 \\
		$\eta=1.0$ & 53.6 & \textbf{84.8}/\textbf{85.9} & \textbf{84.8}/85.6 & 0.537 & 0.765 \\
		$\eta=1.2$ & 53.0 & 83.8/85.8 & 84.0/\textbf{85.7} & 0.541 & \textbf{0.766} \\
		$\eta=1.4$ & 53.6 & 84.1/85.5 & 84.1/85.2 & 0.538 & 0.760 \\
		\midrule[1.5pt]
		$\alpha=0.4$ & 53.6 & 82.4/85.3 & 82.7/85.2 & \textbf{0.532} & \textbf{0.768} \\
		$\alpha=0.6$ & \textbf{53.8} & 82.8/85.5 & 83.1/85.4 & 0.539 & 0.759 \\
		$\alpha=0.8$ & 53.3 & 84.4/85.4 & 84.4/85.2 & 0.542 & 0.765 \\
		$\alpha=1.0$ & 53.6 & \textbf{84.8}/85.9 & \textbf{84.8}/85.6 & 0.537 & 0.765 \\
		$\alpha=1.2$ & \textbf{53.8} & 83.7/85.6 & 83.8/85.4 & 0.536 & 0.762 \\
		$\alpha=1.4$ & 52.3 & \textbf{84.8}/\textbf{86.1} & \textbf{84.8}/\textbf{85.9} & 0.546 & 0.765 \\
		\bottomrule[1.5pt]     
	\end{tabular}
	\caption{Ablation study of MMCL on varying the loss hyperparameters $\mu,\eta,\alpha$ on CMU-MOSEI dataset, which aims at reweighting the contribution of $\mathcal L_{uni}, \mathcal L_{sent}, \mathcal L_{cross}$ in Eq. \ref{equ10}, respectively.}
	\label{table5}
\end{table}



\end{document}